\documentclass{article}
\usepackage{spconf,amsmath,graphicx}
\usepackage{bm}
\usepackage{lipsum}
\usepackage{url}
\usepackage[colorlinks,linkcolor=red]{hyperref}
\usepackage{multirow}
\usepackage{amsfonts,amssymb} 

\title{Silhouette-based View-embeddings for Gait Recognition under Multiple Views}
\name{Tianrui Chai \qquad Xinyu Mei \qquad Annan Li\sthanks{$^*$Corresponding author}\qquad Yunhong Wang}
\address{State Key Laboratory of Virtual Reality Technology and Systems,\\School of Computer Science and Engineering, Beihang University, Beijing 100191, China.\\\{trchai,xymei,liannan,yhwang\}@buaa.edu.cn \thanks{This work was supported by the Key Program of National Natural Science Foundation of China (Grant No. U20B2069). The code is available at \url{https://github.com/ctrasd/gait-view}}}

\begin{document}
%
\maketitle

\begin{abstract}
Gait recognition under multiple views is an important computer vision and pattern recognition task. In the emerging convolutional neural network based approaches, the information of view angle is ignored to some extent. Instead of direct view estimation and training view-specific recognition models, we propose a compatible framework that can embed view information into existing architectures of gait recognition. The embedding is simply achieved by a selective projection layer. Experimental results on two large public datasets show that the proposed framework is very effective. 
\end{abstract}
\vspace{-0.15cm}
\section{Introduction}
\label{sec:intro}

Gait is a biometric presenting the walking style of people and has an edge over other biometrics such as face, fingerprint because it can be recognized at a distance with much less cooperation. Recently, due to the growing demand of intelligent surveillance, gait recognition attracted more attentions.

Variations like carrying conditions, coat-wearing and viewpoint differences may cause changes in gait appearance and bring significant challenges to gait recognition. Solving these problems is of great significance to improve the performance of gait recognition. Among these problems, viewpoint differences is a very tricky problem, because it may bring greater visual differences than the identity.

Over the years many methods have been proposed to solve the problem of multiple view. Gait recognition methods include model-based~\cite{liao2017pose,liao2020model} and appearance-based methods. Appearance-based methods can be divided into two categories, i.e., regarding gait as a single image~\cite{shiraga2016geinet,hu2013view,wu2016comprehensive,he2018multi,li2018beyond,kusakunniran2013recognizing,ben2019coupled,ben2019coupledb} and regarding gait as a video or an image sequence~\cite{chao2019gaitset,fan2020gaitpart,wolf2016multi,lin2020gait,zhang2019cross}. The first category often use gait energy images (GEIs)~\cite{han2005individual} as the representation of gait, while the second category directly makes use of the gait silhouette images. Between these two categories, silhouettes based methods show better performance and become a main trend.

Although the aforementioned methods work well on representing gait in a multi-view scenario, and the deep neural network can somehow learn view-robust feature from mixed views, view itself, i.e. the explicit view estimation and view-specific modeling, is overlook and underrated. We argue that explicit embedding view information can effectively improve the performance of existing approaches.

In this paper, we proposed a general framework for multi-view gait recognition by explicit view angle embedding, based on which, two state-of-the-art gait recognition backbones, i.e.Gaitset~\cite{chao2019gaitset} and GaitGL~\cite{lin2020learning} are enhanced. Compared with the original ones, the enhanced ones, improve the performance. The effectiveness is well demonstrated by the experiments on CASIA-B~\cite{yu2006framework} and OUMVLP~\cite{takemura2018multi} datasets.

The rest of this paper is organized as follows. In Section~\ref{sec:related} we will briefly describe relevant studies. The details of proposed methods are introduced in Section~\ref{sec:method}, while the experimental validation is described om Section~\ref{sec:exp}. And the conclusion is given in Section~\ref{sec:conclude}.

\begin{figure*}[t]
	\centering
	\includegraphics[width=6.5in]{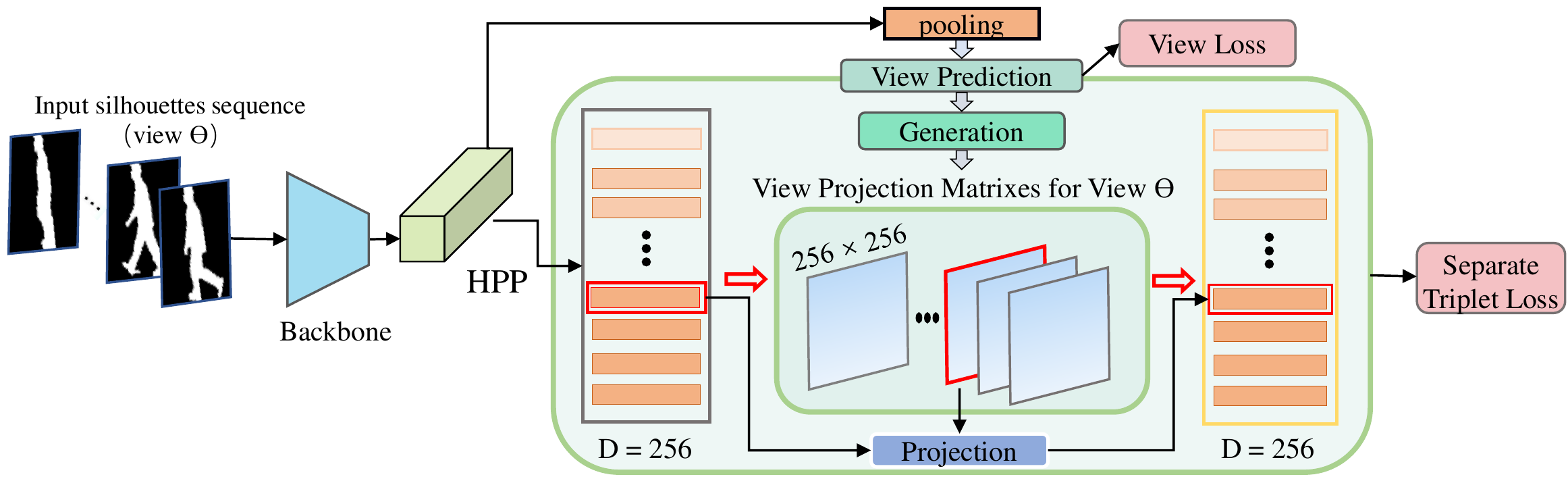}
	\vspace{-0.55cm}
	\caption{Pipeline of our method. The backbone can be replaced by any silhouette-based network, such as Gaitset~\cite{chao2019gaitset}, Gaitpart~\cite{fan2020gaitpart}, MT3D\cite{lin2020gait} and GaitGL~\cite{lin2020learning}. For the extracted feature map, we feed it into two branches. The first branch performs horizontal pyramid pooling (HPP)~\cite{chao2019gaitset} on the feature map. In the second branch, the feature map is pooled to get the view classification features, and the projection matrix is selected according to the predicted view. Then for each feature calculated in HPP at an unique position, we multiply the corresponding projection matrix of the corresponding view to get the final view-robust feature.}
	\label{fig:flowchart}
	\vspace{-0.45cm}
\end{figure*}

\vspace{-0.1cm}
\section{Related work}
\label{sec:related}

\textbf{Gait Recognition}. With the emergence of deep learning, convolution neural networks (CNNs) have been adopted to gait recognition and gain a great success. Shiraga et al.~\cite{shiraga2016geinet} and Wu et al.~\cite{wu2016comprehensive} both used CNN to extract feature from GEIs. However, the simple averaging operation of GEI results in a serious detail loss and limits the learning effects of CNN. To this end, Wolf et al. ~\cite{wolf2016multi} extracted feature directly from original silhouettes by three dimensional convolutions. Chao et al.~\cite{chao2019gaitset} treated gait silhouette sequence as a set and proposed a novel model named Gaitset. Fan et al.~\cite{fan2020gaitpart} further improved silhouette gait feature by introducing part-based representations. Lin et al.~\cite{lin2020gait,lin2020learning} proposed a comprehensive model named GaitGL by integrating both global and local feature. It is recognized as the state-of-the-art gait recognition method.

\noindent\textbf{Multi-task Learning}.
Although success, all the aforementioned approaches overlook the view information which has a great influence on the performance. Multi-task learning, which has been proven effective in many computer vision tasks~\cite{yin2017multi,liang2019multi,dai2016instance}, could be a solution to learning view-specific gait recognition models. This idea is validated by He et al.~\cite{he2018multi}, however, their approach is still based on energy images. To our knowledge, there is no silhouette-based view-specific or aware deep model has been proposed for gait recognition. Besides, multi-task learning is also adopted to gait for recognizing attributes like age and gender~\cite{zhang2019gait,marin2017deep}.

\section{Proposed Method}
\label{sec:method}
What we propose is not a specific model, but a general and compatible framework for multi-view gait recognition. As shown in Figure~\ref{fig:flowchart}, the input is a gait silhouettes sequence $X_{in} \in{\mathbb{R}^{T \times H \times W}}$ and a backbone model $E$ is used to extract feature map $X_{f} \in{\mathbb{R}^{C_f \times H_f \times W_f}}$. The backbone can be any silhouettes-based network, such as Gaitset~\cite{chao2019gaitset}, Gaitpart~\cite{fan2020gaitpart}, MT3D~\cite{liang2019multi} and GaitGL~\cite{lin2020learning}.

Followed by the backbone, the feature map will be fed into two branches. The first one performs Horizontal Pyramid Pooling (HPP)\cite{chao2019gaitset} on the feature map and the Horizontal Pyramid Mapping (HPM) $f_{HPM} \in{\mathbb{R}^{n \times D}}$ will be obtained where $D$ is the dimension of output feature. In the second branch, the feature map is pooled to get the view classification feature $f_v \in{\mathbb{R}^{D_v}} $, and the projection matrices $\{W_1, W_2, W_3,..., W_n\}(W_i\in{\mathbb{R}^{D \times D}})$ are selected according to the predicted view, where $n$ is the number of strips cut in the HPP Module~\cite{chao2019gaitset}. Then for each feature in HPM, we will multiply the projection matrix of the corresponding view to get the final view-invariant feature.

\vspace{-0.2cm}

\subsection{View projection matrix selection}

The feature map $X_f$ is calculated from the input $X_{in}$ using the backbone $E$, and then we use it to compute the view classification feature $f_v$. The process can be expressed as:
\begin{equation}
	X_f=E(X_{in}) 
	~~and~~
	f_v=F(P_{Global\_Avg}(X_f)).
\end{equation}
Especially for Gaitset, another feature map $X_g$ that denotes global feature is calculated from the extractor, then its view classification feature can be defined as:
\begin{equation}
	\begin{aligned}
		f_v&=F(P_{Global\_Avg}([X_f;X_g])),
	\end{aligned}
\end{equation}
where $F$ denotes a fully connect layer and $P_{Global\_Avg}$ denotes global average pooling.

The predicted view probability $\hat{p}\in{\mathbb{R}^M}$ of the input gait silhouettes and the view of maximum probability $\hat{y}$ are calculated as:
\begin{equation}
	\hat{p}=W_{view}f_v+B_{view}
	~~and~~
	\hat{y}=\mathop{\arg\max}_{i}\hat{p_i},
	\label{eq:pro}
	\vspace{-0.2cm}
\end{equation}
where $M$ is the number of discrete views, $W_{view} \in \mathbb{R}^{M\times D_v}$ are weight matrices, $B_{view}$ are the bias terms and $\hat{y} \in \{0,1,2...M\}$.

For predicted view $\hat{y}$, a corresponding view projection matrix group $Z_{\hat{y}}|=\{W_i|i=1,2,...,n\}$ will be trained where $W_i \in \mathbb{R}^{D\times D}$ is the projection matrix. And all the view projection matrix can be expressed as $S=\{Z_i|i=1,2,...,M\}$.

\vspace{-0.1cm}

\subsection{HPP feature projection}

For the convenience for explanation, the horizontal pyramid mapping (HPM) $f_{HPM} \in{\mathbb{R}^{n \times D}}$ is expressed as ${f_{HPM,i}},i=1,2,3,...,n$, where ${f_{HPM,i}}\in{\mathbb{R}^D}$. Suppose that the view $\hat{y}$ of the input gait silhouettes is predicted to be $\theta$ in Equation~\eqref{eq:pro}, then the projected features can be expressed as:
\begin{equation}
	\begin{aligned}
		f_{final,i}&= W_i{f_{HPM,i}}
		\\
		f_{final}&=[f_{final,1},f_{final,2}...,f_{final,n}],
		\label{eq:final_f}
	\end{aligned}
\end{equation}
where $i=1,2,...n$, $W_i \in{Z_{\theta}}$ and the $f_{final}$ is used as the representation for calculating the similarity between two input gait silhouette sequences.

\renewcommand\arraystretch{0.7}
\begin{table*}[t]
	\label{tab:casia}
	\begin{center}
		\caption{Rank-1 accuracy(\%)on CASIA-B under 11 probe views excluding identical-view cases.} 
		\begin{scriptsize}
			\begin{tabular}{|c|c|c|c|c|c|c|c|c|c|c|c|c|c|c|}
				
				\hline
				\multicolumn{3}{|c|}{Gallery NM\#1-4} &
				\multicolumn{12}{c|}{0°-180°} \\ \hline
				\multicolumn{3}{|c|}{Probe} &
				0° &
				18° &
				36° &
				54° &
				72° &
				90° &
				108° &
				128° &
				144° &
				162° &
				180° &
				mean \\ \hline
				\multirow{14}{*}{ST(24)} &
				\multirow{6}{*}{NM\#5-6} &
				ViDP\cite{hu2013view} &
				- &
				- &
				- &
				59.1 &
				- &
				50.2 &
				- &
				57.5 &
				- &
				- &
				- &
				- \\ \cline{3-3}
				&
				&
				CNN-LB\cite{wu2016comprehensive} &
				54.8 &
				- &
				- &
				77.8 &
				- &
				64.9 &
				- &
				76.1 &
				- &
				- &
				- &
				- \\ \cline{3-3}
				&
				&
				GaitSet\cite{chao2019gaitset} &
				64.6 &
				83.3 &
				90.4 &
				86.5 &
				80.2 &
				75.5 &
				80.3 &
				86.0 &
				87.1 &
				81.4 &
				59.6 &
				79.5 \\ \cline{3-3}
				&
				&
				Vi-GaitSet &
				67.8 &
				\textbf{84.3} &
				\textbf{90.7} &
				88.4 &
				\textbf{81.3} &
				\textbf{76.9} &
				82.2 &
				87.2 &
				89.3 &
				84.0 &
				65.8 &
				81.6 \\ \cline{3-3}
				&
				&
				GaitGL\cite{lin2020learning} &
				\textbf{72.1} &
				83.4 &
				88.6 &
				88.6 &
				80.9 &
				74.4 &
				\textbf{82.3} &
				\textbf{88.3} &
				\textbf{89.5} &
				\textbf{86.5} &
				\textbf{69.9} &
				\textbf{82.2} \\ \cline{3-3}
				&
				&
				Vi-GaitGL &
				70.7 &
				83.6 &
				89.0 &
				\textbf{89.1} &
				78.5 &
				71.8 &
				79.6 &
				86.1 &
				88.8 &
				84.7 &
				66.5 &
				80.7 \\ \cline{2-15} 
				&
				\multirow{4}{*}{BG\#1-2} &
				GaitSet\cite{chao2019gaitset} &
				55.8 &
				70.5 &
				76.9 &
				75.5 &
				69.7 &
				63.4 &
				68.0 &
				75.8 &
				76.2 &
				70.7 &
				52.5 &
				68.6 \\ \cline{3-3}
				&
				&
				Vi-GaitSet &
				61.8 &
				74.2 &
				78.9 &
				77.9 &
				\textbf{72.7} &
				\textbf{67.1} &
				71.1 &
				78.6 &
				78.4 &
				71.4 &
				58.8 &
				71.9 \\ \cline{3-3}
				&
				&
				GaitGL\cite{lin2020learning} &
				64.2 &
				73.8 &
				79.3 &
				80.8 &
				71.3 &
				65.3 &
				\textbf{72.3} &
				\textbf{79.2} &
				\textbf{82.5} &
				\textbf{79.7} &
				\textbf{60.3} &
				\textbf{73.5} \\ \cline{3-3}
				&
				&
				Vi-GaitGL &
				\textbf{64.2} &
				\textbf{75.0} &
				\textbf{82.6} &
				\textbf{81.5} &
				70.2 &
				63.9 &
				70.4 &
				77.8 &
				81.0 &
				77.6 &
				58.3 &
				72.9 \\ \cline{2-15} 
				&
				\multirow{4}{*}{CL\#1-2} &
				GaitSet\cite{chao2019gaitset} &
				29.4 &
				43.1 &
				48.5 &
				48.7 &
				42.3 &
				40.3 &
				44.9 &
				47.4 &
				43.0 &
				35.7 &
				25.6 &
				40.9 \\ \cline{3-3}
				&
				&
				Vi-GaitSet &
				33.8 &
				46.4 &
				51.8 &
				47.5 &
				46.8 &
				41.3 &
				44.7 &
				48.5 &
				44.7 &
				36.7 &
				27.5 &
				42.7 \\ \cline{3-3}
				&
				&
				GaitGL\cite{lin2020learning} &
				45.8 &
				59.1 &
				62.7 &
				62.5 &
				56.7 &
				51.5 &
				57.6 &
				60.5 &
				58.0 &
				54.4 &
				37.9 &
				55.1 \\ \cline{3-3}
				&
				&
				Vi-GaitGL &
				\textbf{50.8} &
				\textbf{64.3} &
				\textbf{68.6} &
				\textbf{67.1} &
				\textbf{60.4} &
				\textbf{54.2} &
				\textbf{59.6} &
				\textbf{63.9} &
				\textbf{62.9} &
				\textbf{59.9} &
				\textbf{41.5} &
				\textbf{59.4} \\ \hline
				\multirow{15}{*}{MT(62)} &
				\multirow{5}{*}{NM\#5-6} &
				MGAN\cite{he2018multi} &
				54.9 &
				65.9 &
				72.1 &
				74.8 &
				71.1 &
				65.7 &
				70.0 &
				75.6 &
				76.2 &
				68.6 &
				53.8 &
				68.1 \\ \cline{3-3}
				&
				&
				GaitSet\cite{chao2019gaitset} &
				86.8 &
				95.2 &
				98.0 &
				94.5 &
				91.5 &
				89.1 &
				91.1 &
				95.0 &
				97.4 &
				93.7 &
				80.2 &
				92.0 \\ \cline{3-3}
				&
				&
				Vi-GaitSet &
				87.9 &
				95.8 &
				\textbf{98.5} &
				\textbf{97.1} &
				92.1 &
				89.3 &
				92.5 &
				97.3 &
				97.3 &
				95.4 &
				82.8 &
				93.3 \\ \cline{3-3}
				&
				&
				GaitGL\cite{lin2020learning} &
				90.8 &
				95.3 &
				97.9 &
				96.0 &
				\textbf{94.0} &
				91.4 &
				\textbf{94.7} &
				97.1 &
				\textbf{97.8} &
				\textbf{95.6} &
				\textbf{88.0} &
				\textbf{94.4} \\ \cline{3-3}
				&
				&
				Vi-GaitGL &
				\textbf{90.8} &
				\textbf{95.9} &
				97.7 &
				95.9 &
				93.3 &
				\textbf{91.5} &
				94.4 &
				\textbf{97.3} &
				97.3 &
				95.4 &
				86.9 &
				94.2 \\ \cline{2-15} 
				&
				\multirow{5}{*}{BG\#1-2} &
				MGAN\cite{he2018multi} &
				48.5 &
				58.5 &
				59.7 &
				58.0 &
				53.7 &
				49.8 &
				54.0 &
				51.3 &
				59.5 &
				55.9 &
				43.1 &
				54.7 \\ \cline{3-3}
				&
				&
				GaitSet\cite{chao2019gaitset} &
				79.9 &
				89.8 &
				91.2 &
				86.7 &
				81.6 &
				76.7 &
				81.0 &
				88.2 &
				90.3 &
				88.5 &
				73.0 &
				84.3 \\ \cline{3-3}
				&
				&
				Vi-GaitSet &
				80.8 &
				88.3 &
				92.8 &
				91.3 &
				84.8 &
				79.2 &
				85.3 &
				89.8 &
				92.1 &
				89.9 &
				76.4 &
				86.4 \\ \cline{3-3}
				&
				&
				GaitGL\cite{lin2020learning} &
				\textbf{85.6} &
				\textbf{93.2} &
				\textbf{95.0} &
				92.4 &
				89.0 &
				81.5 &
				86.8 &
				92.7 &
				95.6 &
				92.9 &
				\textbf{83.1} &
				89.8 \\ \cline{3-3}
				&
				&
				Vi-GaitGL &
				83.6 &
				92.9 &
				94.7 &
				\textbf{93.1} &
				\textbf{89.4} &
				\textbf{83.6} &
				\textbf{88.6} &
				\textbf{93.6} &
				\textbf{96.1} &
				\textbf{93.3} &
				81.5 &
				\textbf{90.0} \\ \cline{2-15} 
				&
				\multirow{5}{*}{CL\#1-2} &
				MGAN\cite{he2018multi} &
				23.1 &
				34.5 &
				36.3 &
				33.3 &
				32.9 &
				32.7 &
				34.2 &
				37.6 &
				33.7 &
				26.7 &
				21.0 &
				31.5 \\ \cline{3-3}
				&
				&
				GaitSet\cite{chao2019gaitset} &
				52.0 &
				66.0 &
				72.8 &
				69.3 &
				63.1 &
				61.2 &
				63.5 &
				66.5 &
				67.5 &
				60.0 &
				45.9 &
				62.5 \\ \cline{3-3}
				&
				&
				Vi-GaitSet &
				60.7 &
				70.4 &
				75.8 &
				69.7 &
				65.3 &
				61.3 &
				67.1 &
				70.9 &
				71.9 &
				65.1 &
				49.8 &
				66.2 \\ \cline{3-3}
				&
				&
				GaitGL\cite{lin2020learning} &
				70.2 &
				83.6 &
				87.3 &
				85.2 &
				78.5 &
				73.1 &
				80.0 &
				85.1 &
				84.6 &
				76.9 &
				61.7 &
				78.7 \\ \cline{3-3}
				&
				&
				Vi-GaitGL &
				\textbf{71.2} &
				\textbf{86.5} &
				\textbf{90.9} &
				\textbf{89.0} &
				\textbf{83.9} &
				\textbf{77.2} &
				\textbf{84.8} &
				\textbf{89.1} &
				\textbf{88.6} &
				\textbf{81.0} &
				\textbf{63.7} &
				\textbf{82.3} \\ \hline
				\multirow{18}{*}{LT(74)} &
				\multirow{6}{*}{NM\#5-6} &
				CNN-Ensemble\cite{wu2016comprehensive} &
				88.7 &
				95.1 &
				98.2 &
				96.4 &
				94.1 &
				91.5 &
				93.9 &
				97.5 &
				98.4 &
				95.8 &
				85.6 &
				94.1 \\ \cline{3-3}
				&
				&
				GaitSet\cite{chao2019gaitset} &
				90.8 &
				97.9 &
				\textbf{99.4} &
				96.9 &
				93.6 &
				91.7 &
				95.0 &
				97.8 &
				98.9 &
				96.8 &
				85.8 &
				95.0 \\ \cline{3-3}
				&
				&
				Vi-GaitSet &
				93.1 &
				98.0 &
				99.1 &
				97.1 &
				93.8 &
				92.7 &
				95.8 &
				97.7 &
				99.0 &
				97.7 &
				87.1 &
				95.6 \\ \cline{3-3}
				&
				&
				GaitPart\cite{fan2020gaitpart} &
				94.1 &
				\textbf{98.6} &
				99.3 &
				\textbf{98.5} &
				94.0 &
				92.3 &
				95.9 &
				98.4 &
				\textbf{99.2} &
				97.8 &
				90.4 &
				96.2 \\ \cline{3-3}
				&
				&
				GaitGL\cite{lin2020learning} &
				\textbf{94.6} &
				97.3 &
				98.8 &
				97.1 &
				\textbf{95.8} &
				\textbf{94.3} &
				96.4 &
				98.5 &
				98.6 &
				\textbf{98.2} &
				\textbf{90.8} &
				\textbf{96.4} \\ \cline{3-3}
				&
				&
				Vi-GaitGL &
				93.7 &
				96.9 &
				98.6 &
				97.4 &
				95.5 &
				93.9 &
				\textbf{97.3} &
				\textbf{98.6} &
				98.6 &
				97.7 &
				89.7 &
				96.2 \\ \cline{2-15} 
				&
				\multirow{6}{*}{BG\#1-2} &
				CNN-LB\cite{wu2016comprehensive} &
				64.2 &
				80.6 &
				82.7 &
				76.9 &
				64.8 &
				63.1 &
				68.0 &
				76.9 &
				82.2 &
				75.4 &
				61.3 &
				72.4 \\ \cline{3-3}
				&
				&
				GaitSet\cite{chao2019gaitset} &
				83.8 &
				91.2 &
				91.8 &
				88.8 &
				83.3 &
				81.0 &
				84.1 &
				90.0 &
				92.2 &
				94.4 &
				79.0 &
				87.2 \\ \cline{3-3}
				&
				&
				Vi-GaitSet &
				86.5 &
				93.8 &
				93.4 &
				91.6 &
				86.5 &
				83.0 &
				86.7 &
				91.5 &
				93.2 &
				93.1 &
				81.9 &
				89.2 \\ \cline{3-3}
				&
				&
				GaitPart\cite{fan2020gaitpart} &
				89.1 &
				94.8 &
				96.7 &
				95.1 &
				88.3 &
				84.9 &
				89.0 &
				93.5 &
				96.1 &
				93.8 &
				85.8 &
				91.5 \\ \cline{3-3}
				&
				&
				GaitGL\cite{lin2020learning} &
				\textbf{90.3} &
				\textbf{94.7} &
				\textbf{95.9} &
				94.0 &
				91.9 &
				86.5 &
				90.5 &
				95.5 &
				97.2 &
				\textbf{96.3} &
				\textbf{87.1} &
				92.7 \\ \cline{3-3}
				&
				&
				Vi-GaitGL &
				89.6 &
				94.5 &
				95.6 &
				\textbf{95.2} &
				\textbf{93.2} &
				\textbf{87.3} &
				\textbf{91.7} &
				\textbf{95.9} &
				\textbf{97.8} &
				96.1 &
				85.5 &
				\textbf{92.9} \\ \cline{2-15} 
				&
				\multirow{6}{*}{CL\#1-2} &
				CNN-LB\cite{wu2016comprehensive} &
				37.7 &
				57.2 &
				66.6 &
				61.1 &
				55.2 &
				54.6 &
				55.2 &
				59.1 &
				58.9 &
				48.8 &
				39.4 &
				54.0 \\ \cline{3-3}
				&
				&
				GaitSet\cite{chao2019gaitset} &
				61.4 &
				75.4 &
				80.7 &
				77.3 &
				72.1 &
				70.1 &
				71.5 &
				73.5 &
				73.5 &
				68.4 &
				50.0 &
				70.4 \\ \cline{3-3}
				&
				&
				Vi-GaitSet &
				68.3 &
				78.4 &
				83.4 &
				79.7 &
				72.1 &
				69.5 &
				70.9 &
				75.0 &
				77.7 &
				74.0 &
				58.1 &
				73.4 \\ \cline{3-3}
				&
				&
				GaitPart\cite{fan2020gaitpart} &
				70.7 &
				85.5 &
				86.9 &
				83.3 &
				77.1 &
				72.5 &
				76.9 &
				82.2 &
				83.8 &
				80.2 &
				66.5 &
				78.7 \\ \cline{3-3}
				&
				&
				GaitGL\cite{lin2020learning} &
				76.7 &
				88.3 &
				90.7 &
				86.6 &
				82.7 &
				77.6 &
				83.5 &
				86.5 &
				88.1 &
				83.2 &
				68.7 &
				83.0 \\ \cline{3-3}
				&
				&
				Vi-GaitGL &
				\textbf{81.2} &
				\textbf{92.4} &
				\textbf{94.9} &
				\textbf{93.3} &
				\textbf{87.8} &
				\textbf{82.1} &
				\textbf{87.4} &
				\textbf{89.8} &
				\textbf{90.2} &
				\textbf{87.9} &
				\textbf{72.5} &
				\textbf{87.2} \\ \hline
			\end{tabular}
		\end{scriptsize}
	\end{center}
	\vspace{-0.6cm}
\end{table*}

\subsection{Joint losses}

In the proposed multi-task framework, our loss consists of cross entropy (CE) and triplet loss. Combining the Equation~\eqref{eq:pro}, the CE loss can be expressed as:
\begin{footnotesize} 
	\begin{equation}
		\mathcal{L}_{CE}=-\sum^N_{j=1}\sum^M_{i=1}y_j log(p_{ji})~~w.r.t.~~p_{ji}=\frac{e^{{\hat{p}}_{ji}}}{\sum^M_{i=1}e^{\hat{p}_{ji}}},
		\label{eq:ce}
	\end{equation}
\end{footnotesize}
where $N$ is the number of all gait silhouette sequences and $y_j$ is the discrete ground truth of view of the $j$-th sequence.

Let a triplet of gait silhouette sequences group be $(Q,P,N)$, where $Q$ and $P$ are from the same subject and $Q$ and $N$ are from two different subjects. Denote $K$ triplets of fixed identity as $\{T_i|T_i=(f^{Q_i}_{final},f^{P_i}_{final},f^{N_i}_{final}),i=1,2,...,K\}$, then combining the Equation~\eqref{eq:final_f}, the triplet loss can be expressed as:
\begin{equation}
	\mathcal{L}_{trip}=\frac{1}{K}\sum^K_{i=1}\sum^n_{j=1}max(m-d^{-}_{ij}+d^{+}_{ij},0),
	\label{eq:trip}
\end{equation}
where $d^{-}_{ij}=\|f^{Q_i}_{final,j}-f^{N_i}_{fina,j}\|^2_2$,  $d^{+}_{ij}=\|f^{Q_i}_{final,j}-f^{P_i}_{final,j}\|^2_2$. In this paper we used full mining to make triplets. 

Combine Equation~\eqref{eq:ce} and~\eqref{eq:trip}, the joint loss can be defined as:
\begin{equation}
	\mathcal{L}=\lambda_{CE}\mathcal{L}_{CE}+\lambda_{trip}\mathcal{L}_{trip},
	\label{eq:loss}
\end{equation}
where $\lambda_{CE}$ and $\lambda_{trip}$ are hyper-parameters.

\section{Experiment}
\label{sec:exp}
In order to prove the effectiveness of view embedding in gait recognition, two silhouette-based architectures, i.e. Gaitset~\cite{chao2019gaitset} and GaitGL~\cite{lin2020learning} are used as the backbone.

\vspace{-0.3cm}

\subsection{Datasets}

\textbf{CASIA-B dataset}~\cite{yu2006framework} contains 124 subjects, each contains 11 views and each view contains 10 sequences. The sequences are obtained in three scenarios: normal (NM) (six sequences per subject), walking with bag (BG) (two sequences per subject) and wearing coat or jacket (CL) (two sequences per subject) respectively. We conduct experiments following the settings in~\cite{wu2016comprehensive}. These three settings are small-sample training (ST), medium-sample training (MT) and large-sample training (LT), in which 24, 62 and 74 subjects are used for training and the rest are used for test respectively. The first four sequences of the NM condition (NM\#1-4) are kept in gallery, and the rest sequences are divided into three probe subsets, i.e. NM subsets containing NM \#5-6, BG subsets containing BG \#1-2 and CL subsets containing CL \#1-2. 

\textbf{OU-MVLP dataset}~\cite{takemura2018multi} is the largest public gait dataset, which contains 10,307 subjects. 5,153 subjects are used for training and the rest 5,154 subjects are used for test. Each subject contains 14 views (0°,15°,...,90°;180°,195°,...,270°) and two sequences (\#00-01) per view. In the test set, sequences with index\#01 are kept in gallery and those with index \#00 are used for probes.

\begin{figure}[t]
	\centering
	\includegraphics[width=\linewidth]{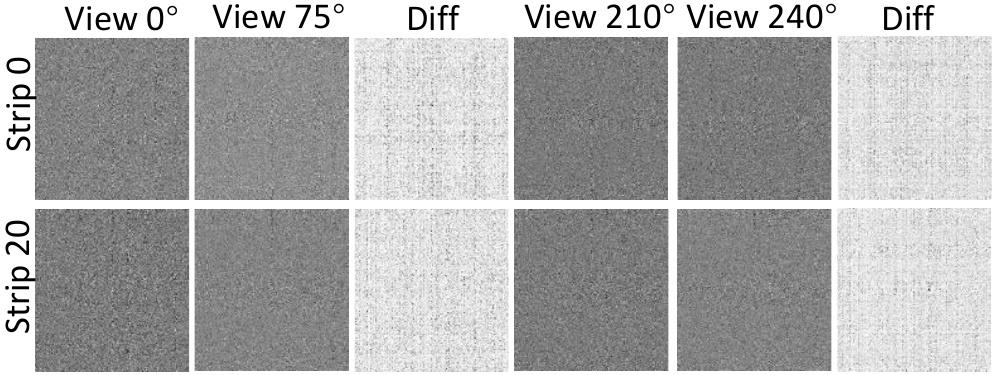}
	\vspace{-0.55cm}
	\caption{Examples of View projection matrices for strip 0 and strip 20. The Diff column shows the absolute difference between the two matrices of different views at the same strip.}
	\label{fig:matrix}
	\vspace{-0.45cm}
\end{figure}

\subsection{Training Details}

In all the experiments, the input is a set of aligned silhouettes of size 64$\times$44 processed by the approach in~\cite{takemura2018multi}. Adam optimizer is used for training and the margin in separate triplet loss is set to 0.2. The $\lambda_{CE}$ and $\lambda_{trip}$ in Equation \eqref{eq:loss} are set to 0.5 and 1.

For backbone Gaitset (Vi-Gaitset), the learning rate is set to 1e-4 for both two datasets. The batch size is set to (8,16) for CASIA-B and (24,16) for OU-MVLP as mentioned in~\cite{chao2019gaitset}. For ST, MT and LT settings of CASIA-B, we train our model for 50K, 60K and 80K iterations. For OU-MVLP, we train our model for 150K iterations.

For GaitGL (Vi-GaitGL), the learning rate is set to 3e-4 for both two datasets at first. The batch size is set to (8,8) for both datasets and the structure of the network is all the same as mentioned in~\cite{lin2020learning}. For ST, MT and LT settings of CASIA-B, we all train our model for 150K iterations and reset the learning rate to 3e-5 for the next 30K iterations. For OU-MVLP, we train our model for 120K iterations. 

Especially, since the sequence number of OU-MVLP is 20 times of CASIA-B, we put our view projection module in different places on the original networks, in order to facilitate the training and prevent overfitting. For CASIA-B, we replace the separate FC module with our view projection module and the parameters of view projection matrices are shared. For OU-MVLP, we add our view projection module after the separate FC module.

Models on CASIA-B are trained with six NVIDIA 2080Ti GPUs and models on OU-MVLP are trained with two Tesla V100-PCIE-32GB GPUs. 
\vspace{-0.4cm}

\subsection{Results and Analysis}

The rank-1 accuracies (\%) on CASIA-B are shown in Table~\ref{tab:casia}. We compared the proposed Vi-Gaitset and Vi-GaitGL with original Gaitset~\cite{chao2019gaitset} and GaitGL~\cite{lin2020learning} as well as VIDP~\cite{hu2013view}, MGAN~\cite{he2018multi} and GaitPart~\cite{fan2020gaitpart}. 

It can be seen that our Vi-Gaitset is more accurate than the original Gaitset under all of the settings by a large margin. For MT and LT settings, the performance of Vi-GaitGL is close to that of the original GaitGL in NM and BG conditions. And the performance of Vi-GaitGL is much better than that of the original GaitGL in CL conditions by 3.6\% and 4.2\%. For ST setting, performance of Vi-GaitGL decreased slightly (1.6\% and 0.6\%) in NM and BG conditions, and increased as high as 4.3\% in CL condition. We argue that part of the performance degradation of Vi-GaitGL in ST setting is due to the small size of training data, which is important to multi-task learning. Another possible explanation is the view recognition accuracy. For Vi-GaitGL, it is only 96.2\% under ST setting, while it is 97.7\% in MT and 97.8\% in LT.

\renewcommand\arraystretch{1.0}
\vspace{-0.4cm} 
\begin{table}[h]
	\setlength{\abovecaptionskip}{0pt}
	\setlength{\belowcaptionskip}{0pt}
	\begin{center}
		\caption{Rank-1 accuracy (\%) on OU-MVLP under 14 probe views excluding identical-view cases.} 
		\vspace{-0.2cm} 
		\label{tab:oumvlp}
		\scalebox{0.72}{
			\begin{tabular}{|c|c|cc|c|cc|}
				\hline
				\multirow{2}{*}{Probe angle} & \multicolumn{6}{c|}{Gallery All 14 views}                                                                                       \\ \cline{2-7} 
				& GEINet & \multicolumn{1}{c}{Gaitset} & Vi-Gaitset    & GaitPart & \multicolumn{1}{c}{GaitGL} & \multicolumn{1}{c|}{Vi-GaitGL} \\ \hline
				0°   & 11.4 & 79.5 & \textbf{81.8} & 82.6 & 84.3 & \textbf{85.6} \\ \cline{1-4} \cline{5-7}
				15°  & 29.1 & 87.9 & \textbf{89.2} & 88.9 & 89.8 & \textbf{90.2} \\ \cline{1-4} \cline{5-7}
				30°  & 41.5 & 89.9 & \textbf{90.5} & 90.8 & 90.8 & \textbf{91.2} \\ \cline{1-4} \cline{5-7}
				45°  & 45.5 & 90.2 & \textbf{90.5} & 91.0 & 91.0 & \textbf{91.5} \\ \cline{1-4} \cline{5-7}
				60°  & 39.5 & 88.1 & \textbf{89.2} & 89.7 & 90.5 & \textbf{91.1} \\ \cline{1-4} \cline{5-7}
				75°  & 41.8 & 88.7 & \textbf{89.5} & 89.7 & 90.5 & \textbf{90.9} \\ \cline{1-4} \cline{5-7}
				90°  & 38.9 & 87.8 & \textbf{89.0} & 89.9 & 90.3 & \textbf{90.4} \\ \cline{1-4} \cline{5-7}
				180° & 14.9 & 81.7 & \textbf{83.9} & 85.2 & 88.1 & \textbf{88.3} \\ \cline{1-4} \cline{5-7}
				195° & 33.1 & 86.7 & \textbf{88.1} & 88.1 & 87.9 & \textbf{88.7} \\ \cline{1-4} \cline{5-7}
				210° & 43.2 & 89.0 & \textbf{89.7} & 90.0 & 89.6 & \textbf{90.6} \\ \cline{1-4} \cline{5-7}
				225° & 45.6 & 89.3 & \textbf{89.8} & 90.1 & 89.8 & \textbf{90.6} \\ \cline{1-4} \cline{5-7}
				240° & 39.4 & 87.2 & \textbf{88.6} & 89.0 & 88.9 & \textbf{90.1} \\ \cline{1-4} \cline{5-7}
				255° & 40.5 & 87.8 & \textbf{88.5} & 89.1 & 88.9 & \textbf{89.9} \\ \cline{1-4} \cline{5-7}
				270° & 36.3 & 86.2 & \textbf{87.6} & 88.2 & 88.2 & \textbf{89.4} \\ \cline{1-7}
				mean                         & 35.8   & \multicolumn{1}{c}{87.1}    & \textbf{88.3} & 88.7     & 89.1                        & \textbf{89.9}                  \\ \cline{1-7}												
		\end{tabular}}
	\end{center}
	\vspace{-0.4cm}
\end{table}
\vspace{-0.1cm}

The results in rank-1 accuracy (\%) on OU-MVLP dataset are shown in Table~\ref{tab:oumvlp}. The performance of Vi-Gaitset is better than the original Gaitset~\cite{chao2019gaitset} under all the probe views. The proposed Vi-GaitGL meets a new state-of-the-art under various cross-view conditions and the mean rank-1 accuracy is 0.8\% higher than the original GaitGL~\cite{lin2020learning}. When the data is sufficient, out method is consistently better.

In order to explain the effectiveness of our framework, we compare the projection matrices of different views in Vi-GaitGL (trained on OU-MVLP). As illustrated in Figure \ref{fig:matrix}, their difference has obvious vertical texture, which indicates that the projection matrices of different views has view specificity for feature mapping.

\vspace{-0.3cm}

\section{Conclusion}
\label{sec:conclude}

\vspace{-0.3cm}
In this paper, we propose a general view embedding framework for improved multi-view gait recognition, in which the view angle is explicitly estimated and used for model refining. Experimental results on two leading backbone models show that our idea of explicit view embedding is very effective. The proposed framework with GaitGL~\cite{lin2020learning} as the backbone meets the state-of-the-art on two large-scale public gait datasets. It should be pointed out that the proposed framework is not competitive but rather complementary to existing works.

\newpage

\bibliographystyle{IEEEbib}
\bibliography{ref}

\begin{thebibliography}{10}

\bibitem{shiraga2016geinet}
Kohei Shiraga, Yasushi Makihara, Daigo Muramatsu, Tomio Echigo, and Yasushi
  Yagi,
\newblock ``Geinet: View-invariant gait recognition using a convolutional
  neural network,''
\newblock in {\em ICB}. IEEE, 2016, pp. 1--8.

\bibitem{hu2013view}
Maodi Hu, Yunhong Wang, Zhaoxiang Zhang, James~J Little, and Di~Huang,
\newblock ``View-invariant discriminative projection for multi-view gait-based
  human identification,''
\newblock {\em IEEE TIFS}, vol. 8, no. 12, pp. 2034--2045, 2013.

\bibitem{wu2016comprehensive}
Zifeng Wu, Yongzhen Huang, Liang Wang, Xiaogang Wang, and Tieniu Tan,
\newblock ``A comprehensive study on cross-view gait based human identification
  with deep cnns,''
\newblock {\em IEEE TPAMI}, vol. 39, no. 2, pp. 209--226, 2016.

\bibitem{chao2019gaitset}
Hanqing Chao, Yiwei He, Junping Zhang, and Jianfeng Feng,
\newblock ``Gaitset: Regarding gait as a set for cross-view gait recognition,''
\newblock in {\em AAAI}, 2019, vol.~33, pp. 8126--8133.

\bibitem{fan2020gaitpart}
Chao Fan, Yunjie Peng, Chunshui Cao, Xu~Liu, Saihui Hou, Jiannan Chi, Yongzhen
  Huang, Qing Li, and Zhiqiang He,
\newblock ``Gaitpart: Temporal part-based model for gait recognition,''
\newblock in {\em CVPR}, 2020, pp. 14225--14233.

\bibitem{lin2020gait}
Beibei Lin, Shunli Zhang, and Feng Bao,
\newblock ``Gait recognition with multiple-temporal-scale 3d convolutional
  neural network,''
\newblock in {\em ACM MM}, 2020, pp. 3054--3062.

\bibitem{wolf2016multi}
Thomas Wolf, Mohammadreza Babaee, and Gerhard Rigoll,
\newblock ``Multi-view gait recognition using 3d convolutional neural
  networks,''
\newblock in {\em ICIP}. IEEE, 2016.

\bibitem{han2005individual}
Jinguang Han and Bir Bhanu,
\newblock ``Individual recognition using gait energy image,''
\newblock {\em TPAMI}, vol. 28, no. 2, pp. 316--322, 2005.

\bibitem{he2018multi}
Yiwei He, Junping Zhang, Hongming Shan, and Liang Wang,
\newblock ``Multi-task gans for view-specific feature learning in gait
  recognition,''
\newblock {\em IEEE TIFS}, vol. 14, no. 1, pp. 102--113, 2018.

\bibitem{zhang2019cross}
Yuqi Zhang, Yongzhen Huang, Shiqi Yu, and Liang Wang,
\newblock ``Cross-view gait recognition by discriminative feature learning,''
\newblock {\em IEEE TIP}, vol. 29, pp. 1001--1015, 2019.

\bibitem{yu2006framework}
Shiqi Yu, Daoliang Tan, and Tieniu Tan,
\newblock ``A framework for evaluating the effect of view angle, clothing and
  carrying condition on gait recognition,''
\newblock in {\em ICPR}. IEEE, 2006, vol.~4, pp. 441--444.

\bibitem{takemura2018multi}
Noriko Takemura, Yasushi Makihara, Daigo Muramatsu, Tomio Echigo, and Yasushi
  Yagi,
\newblock ``Multi-view large population gait dataset and its performance
  evaluation for cross-view gait recognition,''
\newblock {\em IPSJ TCVA}, vol. 10, no. 1, pp. 4, 2018.

\bibitem{li2018beyond}
Shuangqun Li, Wu~Liu, Huadong Ma, and Shaopeng Zhu,
\newblock ``Beyond view transformation: Cycle-consistent global and partial
  perception gan for view-invariant gait recognition,''
\newblock in {\em ICME}, 2018.

\bibitem{lin2020learning}
Beibei Lin, Shunli Zhang, Xin Yu, Zedong Chu, and Haikun Zhang,
\newblock ``Learning effective representations from global and local features
  for cross-view gait recognition,''
\newblock {\em arXiv preprint arXiv:2011.01461}, 2020.

\bibitem{marin2017deep}
Manuel~J Mar{\'\i}n-Jim{\'e}nez, Francisco~M Castro, Nicol{\'a}s Guil,
  F~de~La~Torre, and R~Medina-Carnicer,
\newblock ``Deep multi-task learning for gait-based biometrics,''
\newblock in {\em ICIP}. IEEE, 2017, pp. 106--110.

\bibitem{liang2019multi}
Ming Liang, Bin Yang, Yun Chen, Rui Hu, and Raquel Urtasun,
\newblock ``Multi-task multi-sensor fusion for 3d object detection,''
\newblock in {\em CVPR}, 2019, pp. 7345--7353.

\bibitem{yin2017multi}
Xi~Yin and Xiaoming Liu,
\newblock ``Multi-task convolutional neural network for pose-invariant face
  recognition,''
\newblock {\em IEEE TIP}, vol. 27, no. 2, pp. 964--975, 2017.

\bibitem{dai2016instance}
Jifeng Dai, Kaiming He, and Jian Sun,
\newblock ``Instance-aware semantic segmentation via multi-task network
  cascades,''
\newblock in {\em CVPR}, 2016, pp. 3150--3158.

\bibitem{zhang2019gait}
Shaoxiong Zhang, Yunhong Wang, and Annan Li,
\newblock ``Gait-based age estimation with deep convolutional neural network,''
\newblock in {\em ICB}. IEEE, 2019, pp. 1--8.

\bibitem{kusakunniran2013recognizing}
Worapan Kusakunniran, Qiang Wu, Jian Zhang, Hongdong Li, and Liang Wang,
\newblock ``Recognizing gaits across views through correlated motion
  co-clustering,''
\newblock {\em IEEE TIP}, vol. 23, no. 2, pp. 696--709, 2013.

\bibitem{liao2017pose}
Rijun Liao, Chunshui Cao, Edel~B Garcia, Shiqi Yu, and Yongzhen Huang,
\newblock ``Pose-based temporal-spatial network (ptsn) for gait recognition
  with carrying and clothing variations,''
\newblock in {\em CCBR}. Springer, 2017, pp. 474--483.

\bibitem{liao2020model}
Rijun Liao, Shiqi Yu, Weizhi An, and Yongzhen Huang,
\newblock ``A model-based gait recognition method with body pose and human
  prior knowledge,''
\newblock {\em Pattern Recognition}, vol. 98, pp. 107069, 2020.

\bibitem{ben2019coupled}
Xianye Ben, Chen Gong, Peng Zhang, Xitong Jia, Qiang Wu, and Weixiao Meng,
\newblock ``Coupled patch alignment for matching cross-view gaits,''
\newblock {\em IEEE TIP}, vol. 28, no. 6, pp. 3142--3157, 2019.

\bibitem{ben2019coupledb}
Xianye Ben, Chen Gong, Peng Zhang, Rui Yan, Qiang Wu, and Weixiao Meng,
\newblock ``Coupled bilinear discriminant projection for cross-view gait
  recognition,''
\newblock {\em IEEE TCSVT}, vol. 30, no. 3, pp. 734--747, 2019.

\end{thebibliography}

\end{document}